\def\year{2021}\relax
\documentclass[letterpaper]{article} 
\usepackage{aaai21}  
\usepackage{times}  
\usepackage{helvet} 
\usepackage{courier}  
\usepackage[hyphens]{url}  
\usepackage{graphicx} 
\usepackage{natbib}  
\usepackage{caption} 
\usepackage{adjustbox}
\usepackage{latexsym}
\usepackage{multirow}
\usepackage{diagbox}
\usepackage{amsmath}
\usepackage{pifont}
\usepackage{amssymb}
\usepackage{booktabs}
\usepackage{makecell}
\usepackage{subfigure}
\usepackage[ruled]{algorithm2e} 
\usepackage{microtype}
\usepackage{todonotes}
\usepackage{kotex}

\newcommand{\cls}{[\textsc{cls}]}
\newcommand{\sep}{[\textsc{sep}]}
\newcommand{\ins}{[\textsc{ins}]}
\newcommand{\del}{[\textsc{del}]}
\newcommand{\srch}{[\textsc{srch}]}
\newcommand{\eot}{[\textsc{eot}]}
\title{AAAI Press Formatting Instructions \\for Authors Using \LaTeX{} --- A Guide }
\author{

    Written by AAAI Press Staff\textsuperscript{\rm 1}\thanks{With help from the AAAI Publications Committee.}\\
    AAAI Style Contributions by Pater Patel Schneider,
    Sunil Issar,  \\
    J. Scott Penberthy,
    George Ferguson,
    Hans Guesgen,
    Francisco Cruz,
    Marc Pujol-Gonzalez
    \\
}
\affiliations{
    \textsuperscript{\rm 1}Wisenut Inc. \\
    \textsuperscript{\rm 2}Kakao Corp. \\
    \textsuperscript{\rm 3}Korea University \\
    \textsuperscript{\rm 4}Kakao Enterprise Corp. \\
    
}

\makeatletter
\newcommand{\printfnsymbol}[1]{%
  \textsuperscript{\@fnsymbol{#1}}%
}

\title{Do Response Selection Models Really Know What's Next? \\ Utterance Manipulation Strategies For Multi-turn Response Selection}
\author {
    Taesun Whang\textsuperscript{\rm 1}\thanks{These authors equally contributed to this work.}\ \ ~Dongyub Lee\textsuperscript{\rm 2}\printfnsymbol{1}\ \ ~Dongsuk Oh\textsuperscript{\rm 3}\ \ ~Chanhee Lee\textsuperscript{\rm 3}\\ ~Kijong Han\textsuperscript{\rm 4}\ \ ~Dong-hun Lee\textsuperscript{\rm 4}\ \ ~Saebyeok Lee\textsuperscript{\rm 1,3}\hspace{0.005em}\thanks{Corresponding author.}\\
}

\begin{document}

\maketitle

\begin{abstract}
In this paper, we study the task of selecting the optimal response given a user and system utterance history in retrieval-based multi-turn dialog systems. Recently, pre-trained language models ({\em{e.g.,}} BERT, RoBERTa, and ELECTRA) showed significant improvements in various natural language processing tasks. This and similar response selection tasks can also be solved using such language models by formulating the tasks as dialog--response binary classification tasks. Although existing works using this approach successfully obtained state-of-the-art results, we observe that language models trained in this manner tend to make predictions based on the relatedness of history and candidates, ignoring the sequential nature of multi-turn dialog systems. This suggests that the response selection task alone is insufficient for learning temporal dependencies between utterances. To this end, we propose utterance manipulation strategies (UMS) to address this problem. Specifically, UMS consist of several strategies ({\em{i.e.,}} insertion, deletion, and search), which aid the response selection model towards maintaining dialog coherence. 
Further, UMS are self-supervised methods that do not require additional annotation and thus can be easily incorporated into existing approaches. Extensive evaluation across multiple languages and models shows that UMS are highly effective in teaching dialog consistency, which leads to models pushing the state-of-the-art with significant margins on multiple public benchmark datasets. 
\end{abstract}
\section{Introduction}

\begin{figure}[t]\centering
\includegraphics[width=0.4\textwidth]{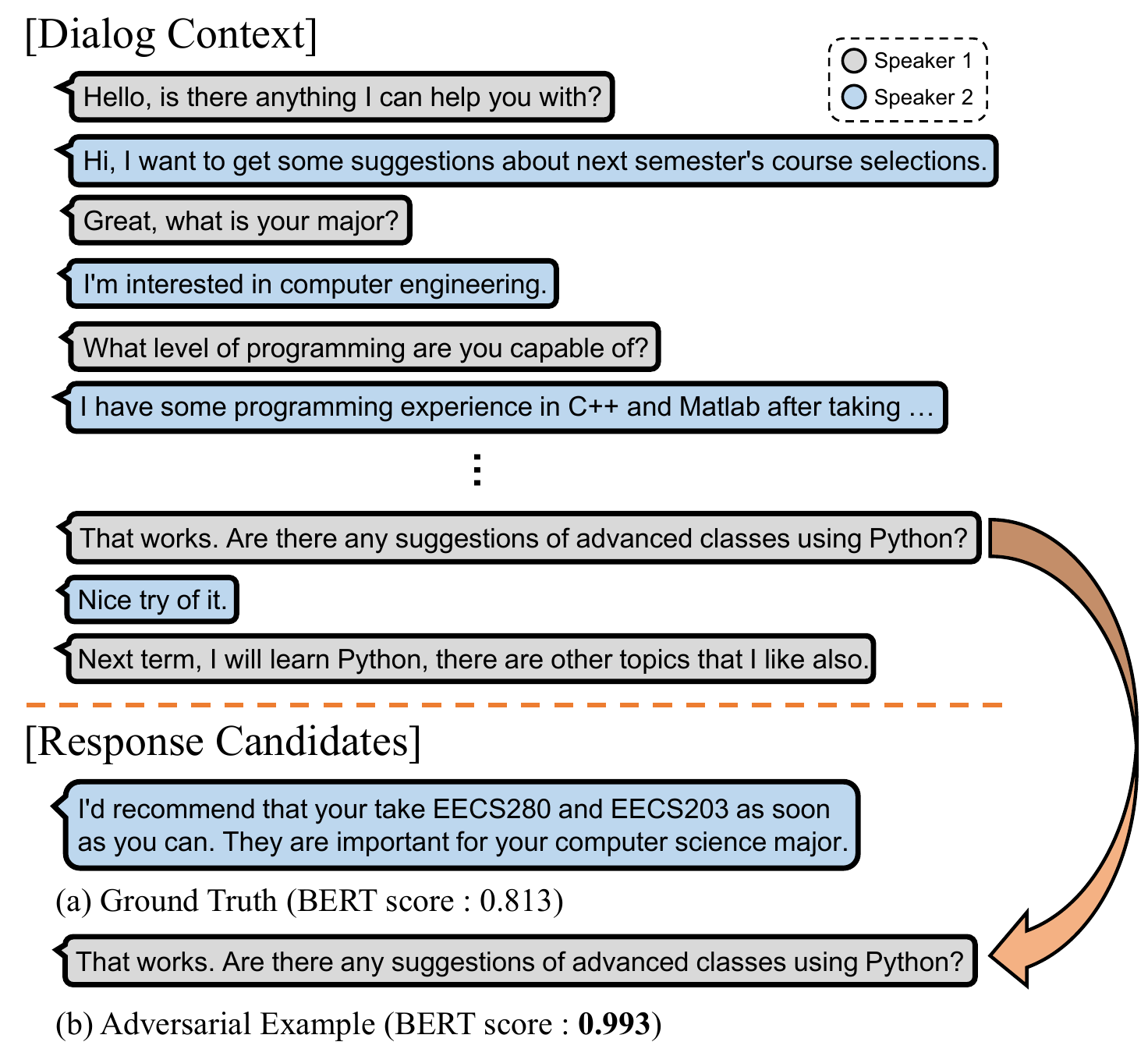}
\caption{Example of multi-turn response selection. BERT-based model tends to calculate the matching score of a dialog--response pair depending on the semantic relatedness of the dialog and the response ((a) $<$ (b)). More details are in Discussion section.}
\label{fig:intro_example}
\end{figure}

In recent years, building intelligent conversational agents has gained considerable attention in the field of natural language processing (NLP). Among widely used dialog systems, retrieval-based dialog systems \cite{lowe2015ubuntu,wu2017sequential,zhang2018dua} are implemented in a variety of industries because they provide accurate, informative, and promising responses. In this study, we focus on multi-turn response selection in retrieval-based dialog systems. This is a task of predicting the most likely response under a given dialog history from a set of candidates.

Existing works \cite{wu2017sequential,zhou2018multi,tao2019multi,yuan2019multi} have studied utterance--response matching based on attention mechanisms including self-attention \cite{vaswani2017attention}. Most recently, as pre-trained language models ({\em{e.g.,}} BERT \cite{devlin2018bert}, RoBERTa \cite{liu2019roberta}, and ELECTRA \cite{clark2020electra}) have achieved substantial performance improvements in diverse NLP tasks, multi-turn response selection also has been resolved using such language models \cite{whang2019domain,lu2020improving,gu2020sabert,humeau2019poly}.

However, we tackle three crucial problems in applying language models to response selection. 1) Domain adaptation based on an additional training on a target corpus is extremely time-consuming and computationally costly. 2) Formulating response selection as a dialog--response binary classification task is insufficient to represent intra- and inter-utterance interactions as the dialog context is formed by concatenating all utterances. 3) The models tend to select the optimal response depending on how semantically similar it is to a given dialog. 
As shown in Figure~\ref{fig:intro_example}, we experiment to verify whether the BERT-based response selection model is trained properly to select the next utterance rather than dialog-related response. The results show that the model tends to give a higher probability score to a response that is more semantically related to the dialog context rather than consistent response. Although it is obvious that the ground truth is suitable for being the next utterance, the model highly depends on its semantic meaning.

To address these issues, this paper proposes Utterance Manipulation Strategies (UMS) for multi-turn response selection. Specifically, UMS consist of three powerful strategies ({\em{i.e.,}} insertion, deletion, and search), which effectively help the response selection model to learn temporal dependencies between utterances as well as semantic matching and maintain dialog coherence. In addition, these strategies are fully self-supervised methods that do not require additional annotation and can be easily adapted to existing studies. We briefly summarize the main contributions of this paper as follows: 1) We show that existing response selection models are more likely to predict a semantically relevant response with its dialog rather than the next utterance. 2) We propose simple but novel utterance manipulation strategies, which are highly effective in predicting the next utterance. Our model has strengths in effectively performing in-domain classification. 3) Experimental results on three benchmarks ({\em{i.e.,}} Ubuntu, Douban, and E-commerce) show that our proposed model outperforms state-of-the-art methods. We also obtain significant improvements in performance compared to the baselines on a new Korean open-domain corpus.
\section{Proposed Method}
\subsection{Language Models for Response Selection}
\noindent\textbf{Pre-trained Language Models} Recently, pre-trained language models, such as BERT \cite{devlin2018bert} and ELECTRA \cite{clark2020electra}, were successfully adapted to a wide range of NLP tasks, including multi-turn response selection, achieving state-of-the-art results. In this work, we build upon this success and evaluate our method by incorporating it into BERT and ELECTRA.\\
\noindent\textbf{Domain-specific Post-training} As contextual language models are pre-trained on general corpora, such as the Toronto Books Corpus and Wikipedia, it is less effective to directly fine-tune these models on downstream tasks if there is a domain shift. Hence, it is a common practice to further train such models with the language modeling objective using texts from the target domain to reduce the negative impact. This has shown to be effective in various tasks including review reading comprehension \cite{xu2019bert} and SuperGLUE \cite{wang2019superglue}. Existing works on multi-turn response selection \cite{whang2019domain,gu2020sabert,humeau2019poly} also adapted this post-training approach and obtained state-of-the-art results. We also employ this post-training method in this work and show its effectiveness in improving performance.\\
\noindent\textbf{Training Response Selection Models} Following several researches based on contextual language models for multi-turn response selection \cite{whang2019domain,lu2020improving,gu2020sabert}, a pointwise approach is used to learn a cross-encoder that receives both dialog context and response simultaneously. Suppose that a dialog agent is given a dialog dataset $\mathcal{D}=\{(U_i,r_i,y_i)\}_{i=1}^{N}$. Each triplet consists of 1) a sequence of utterances $U_i = [u^i_1, u^i_2, \cdots, u^i_{|U|}] $ representing the historical context, where $u^i_t$ is a single utterance, 2) a response $r_i$, and 3) a label $y_i \in \{0,1\}$. Each utterance $u^i_t$ and response $r_i$ are composed of multiple tokens including a special \eot\, token at the end of each utterance, following the work of \citet{whang2019domain}. In general, the input sequence,
\begin{align*}\label{eq:ressel_input}
\mathbf{X} = [\cls\,u_1\,u_2\,...\,u_{n_u}\,\sep\,r\,\sep],
\end{align*}
is fed into pre-trained language models ({\em{i.e.,}} BERT, ELECTRA), then output representation of \cls\, token, $\mathbf{x}_{\cls}\in\mathbb{R}^{d\times1}$, is used to classify whether the dialog--response pair is consistent. The relevance score of the dialog utterances and response is formulated as, 
\begin{equation}\label{eq:ressel_loss}
g(U,r) = \sigma(\mathbf{w}^{\top}\mathbf{x}_{\cls}+b),
\end{equation}
where $\mathbf{w}\in\mathbb{R}^{d\times1}$ and $b$ are the trainable parameters. We use binary cross-entropy loss to optimize the models.

\begin{figure*}[t]\centering
\includegraphics[width=0.9\textwidth]{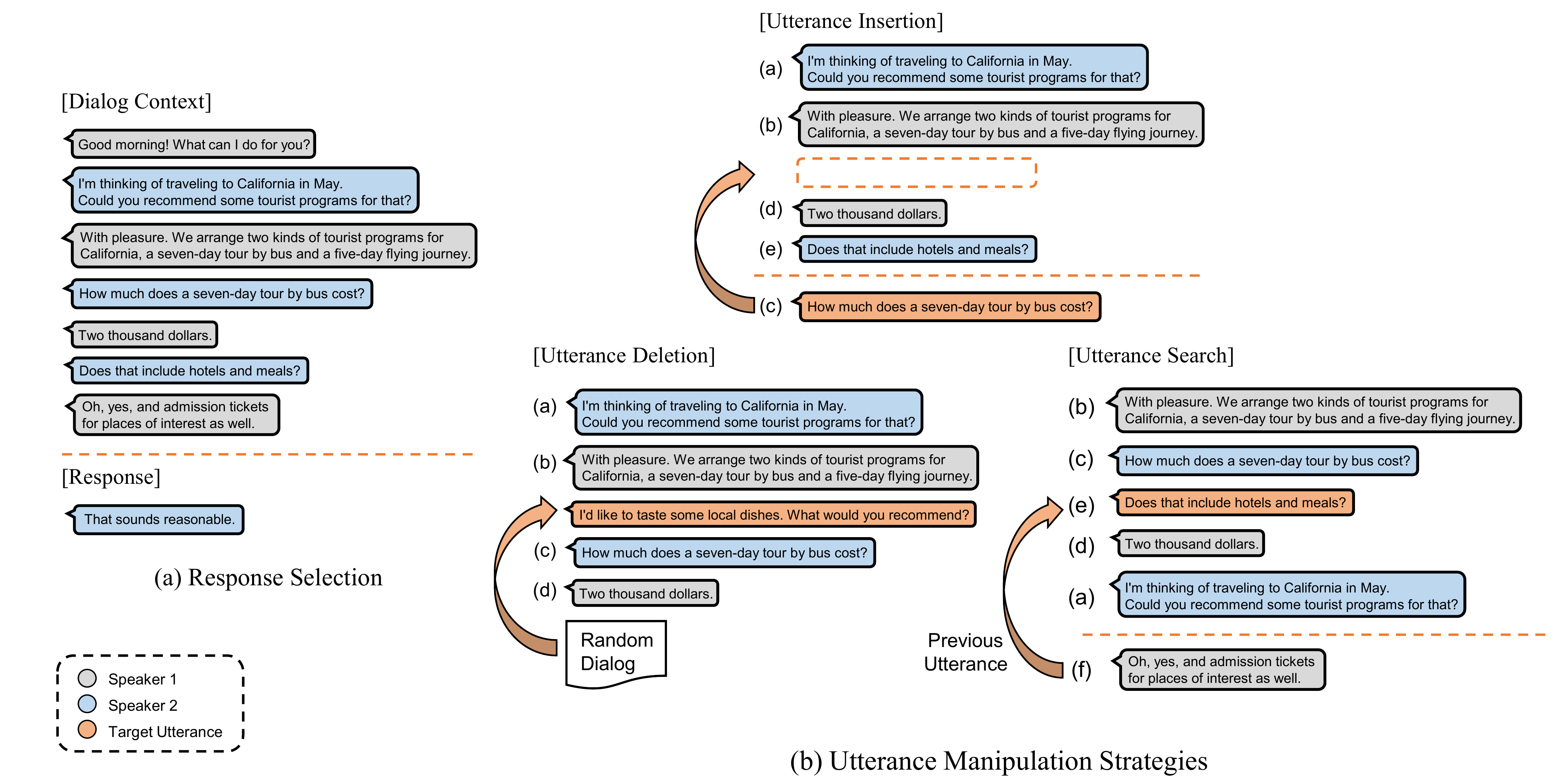}
\caption{An overview of Utterance Manipulation Strategies. Input sequence for each manipulation strategy is dynamically constructed by extracting $k$ consecutive utterances from the original dialog context during the training period. Also, target utterance is randomly chosen from either the dialog context (Insertion, Search) or the random dialog (Deletion).}
\label{fig:model_overview}
\end{figure*}

\subsection{Utterance Manipulation Strategies}
Figure \ref{fig:model_overview} describes the overview of our proposed method, utterance manipulation strategies. We propose a multi-task learning framework, which consists of three highly effective auxiliary tasks for multi-turn response selection, utterance 1) \textit{insertion}, 2) \textit{deletion}, and 3) \textit{search}.  These tasks are jointly trained with the response selection model during the fine-tuning period. To train the auxiliary tasks, we add new special tokens, \ins, \del, and \srch\, for the utterance insertion, deletion, and search tasks, respectively. We cover how we train the model with these special tokens in the following sections.\\
\noindent\textbf{Utterance Insertion} 
Despite the huge success of BERT, it has limitations in understanding discourse-level semantic structure since NSP, one of BERT's objectives, mainly learns semantic topic shift rather than sentence order \cite{lan2020albert}.
In multi-turn response selection, the model needs the ability not only to distinguish the utterances with different semantic meanings but also to discriminate whether the utterances are consecutive even if they are semantically related. We propose \textit{utterance insertion} to resolve the aforementioned issues.\\
\indent We first extract $k$ consecutive utterances from the original dialog context, then randomly select one of the utterances to be inserted. To train the model to find where the selected utterance should be inserted, \ins\, tokens are positioned before each utterance and after the last utterance. \ins\, tokens are represented as possible position of the target utterance. Input sequence for utterance insertion is denoted as
\begin{align*}
\mathbf{X}_{\textsc{ins}} = [\cls\,\ins_1\,u_1\ins_2\,u_2\,...\,u_{t-1}\qquad\\
\ins_{t}\,u_{t+1}\,...\,u_{k}\,\ins_k\,\sep\,u_{t}\,\sep]\text{,}
\end{align*}
where $u_{t}$ is the target utterance and \ins$_{t}$ is the target insertion token.\\
\noindent\textbf{Utterance Deletion}
Recent BERT-based models for multi-turn response selection regard the task as a dialog--response binary classification. Even though they are extended in a multi-turn manner using separating tokens ({\em{e.g.,}} \sep, \eot\,), 
these models lack utterance-level interaction between dialog context and response. 
To alleviate this, we propose a highly effective auxiliary task, \textit{utterance deletion}, to enrich utterance-level interaction in multi-turn conversation. \\
\indent As with utterance insertion, $k$ consecutive utterances are extracted from the original dialog context, and then an utterance from a random dialog is inserted among the $k$ extracted utterances. In other words, $k+1$ utterances are composed of $k$ utterances from the original conversation and one from different dialogs. To train the model to find an unrelated utterance, \del\, tokens are positioned
before each utterance. The objective of the utterance deletion task is to predict which utterance causes inconsistency. We denote the input sequence for utterance deletion as
\begin{align*}
\mathbf{X}_{\textsc{del}} = [\cls\,\del_1\,u_1\,\del_2\,u_2\,...\del_{t}\qquad\\
\,u^{rand}\,\del_{t+1}\,u_{t}\,...\,\del_{k+1}\,u_{k}\,\sep]\text{,}
\end{align*}
where $u^{rand}$ is the utterance from the random dialog and $\del_{t}$ is the target deletion token.\\
\noindent\textbf{Utterance Search}
Whereas two previous auxiliary tasks are performed in a properly ordered dialog, we design a novel task, \textit{utterance search}, which aims to find an appropriate utterance from randomly shuffled utterances. The objective of this task is to learn temporal dependencies between semantically similar utterances.\\
\indent Given $k$ consecutive utterances same as the previous tasks, we shuffle utterances except for the last utterance and insert \srch\, tokens before each shuffled utterance. The utterance search aims to find the previous utterance of the last utterance from the jumbled utterances. Input sequence for utterance search is denoted as
\begin{align*}
\mathbf{X}_{\textsc{srch}} = [\cls\,\srch_1\,u'_1\srch_2\,u'_2\,...\qquad\\
\srch_{t}\,u'_{t}\,...\,u'_{k-1}\sep\,u_{k}\,\sep]\text{,}
\end{align*}
where $\{u'_{t}\}_{t=1}^{k-1}$ is a set of utterances which are randomly shuffled except for the last utterance $u_k$. The previous utterance of $u_k$ is denoted as $u'_t$ ({\em{i.e.,}} $u_{k-1}$) and \srch$_{t}$ is the target search token.

\subsection{Multi-Task Learning Setup}
The input sequence of each task is fed into the language models. The output representations of special tokens ({\em{i.e.,}} \ins, \del, and \srch\,) are used to classify whether each token is in a correct position to be inserted, deleted, and searched. Target tokens for each task ({\em{i.e.,}} \ins$_{t}$,\,\del$_{t}$, and \srch$_{t}$) are labeled as 1, otherwise 0. We calculate the probability of the token being a target as follows:
\begin{equation}
p(y_{\textsc{task}}=1|\mathbf{X}_{\textsc{task}}) = \sigma(\mathbf{w}^{\top}\mathbf{x}_{\textsc{task}}+b),
\end{equation}
where \textsc{task}$\in$\{\textsc{ins},\,\textsc{del},\,\textsc{srch}\} and $\mathbf{x}_\textsc{task}$ is the output representations of each special token. We use binary cross-entropy loss for all auxiliary tasks to optimize each model. The final loss is determined by summing the response selection loss and UMS losses with the same ratio.
\begin{table*}[t]\centering
\begin{adjustbox}{width=0.75\textwidth}

\begin{tabular}{l | ccc | ccc | ccc | cccc}
\toprule
\multirow{2}{*}{Dataset} & \multicolumn{3}{c|}{Ubuntu}  & \multicolumn{3}{c|}{Douban}  & \multicolumn{3}{c|}{E-Commerce} & \multicolumn{4}{c}{Kakao}\\
                         & Train & Val   & Test  & Train & Val  & Test       & Train & Val  & Test  & Train & Val  & Test (Web) & Test (Clean)\\
\midrule
\# pairs                 & 1M    & 500K  & 500K  & 1M    & 50K  & 6670       & 1M    & 10K  & 10K & 1M & 50K & 5139 & 7164 \\
pos:neg                  & 1:1   & 1:9   & 1:9   & 1:1   & 1:1  & 1.2:8.8    & 1:1   & 1:1  & 1:9 & 1:1 & 1:1 & 1.6:7.4 & 2:7 \\
\# avg turns             & 10.13 & 10.11 & 10.11 & 6.69  & 6.75 & 6.45     & 5.51  & 5.48 & 5.64 & 3.00 & 3.00 & 3.49 & 3.25 \\
\bottomrule

\end{tabular}
\end{adjustbox}

\caption{Corpus statistics of multi-turn response selection datasets.}
\label{table:corpus_stats}
\end{table*}

\section{Experimental Setup}
\subsection{Datasets}
We evaluate our model on three widely used response selection benchmarks: \textit{Ubuntu Corpus V1} \cite{lowe2015ubuntu}, \textit{Douban Corpus} \cite{wu2017sequential}, and \textit{E-Commerce Corpus} \cite{zhang2018dua}. In addition, a new open-domain dialog corpus, \textit{Kakao Corpus}, is utilized to evaluate our model. All datasets consist of dyadic multi-turn conversations, and their statistics are summarized in Table \ref{table:corpus_stats}.\\
\noindent\textbf{Ubuntu Corpus V1} Ubuntu dataset is a large multi-turn conversation corpus that is constructed from Ubuntu internet relay chats. It mainly consists of conversations between two participants who discuss how to troubleshoot the Ubuntu operating system. We utilize the data released by \citet{xu2017incorporating}, where numbers, URLs, paths are replaced with special placeholders following previous works \cite{wu2017sequential, zhou2018multi}.\\
\noindent\textbf{Douban Corpus} Douban dataset is a Chinese open-domain dialog corpus, whereas the Ubuntu Corpus is a domain-specific dataset. It is constructed by web-crawling from the Douban group\footnote{\begin{footnotesize}https://www.douban.com\end{footnotesize}}, which is a popular social networking service (SNS) in China.\\
\noindent\textbf{E-commerce Corpus} E-Commerce dataset is another Chinese multi-turn conversation corpus. It is collected from real-world customer consultation dialogs from Taobao\footnote{\begin{footnotesize}https://www.taobao.com\end{footnotesize}}, which is the largest Chinese e-commerce platform. It consists of several types of conversations ({\em{e.g.,}} commodity consultation, recommendation, and negotiation) based on various commodities.\\
\noindent\textbf{Kakao Corpus} Kakao dataset is a large Korean open-domain dialog corpus that is constructed by Kakao Corporation\footnote{\begin{footnotesize}https://www.kakaocorp.com\end{footnotesize}}. It is mainly web-crawled from Korean SNSs such as Korean Twitter and Reddit. In a similar manner to the Ubuntu dataset, we take the last utterance of the dialog as a positive response and the rest as a dialog context. Negative responses are randomly sampled from the other conversations. We split the test set into two sets: 1) \textit{web} is the same as the training set, and 2) \textit{clean} consists of grammatically correct conversations that are constructed by human annotators and inspected by NLP experts.

\begin{table*}[t]
\centering
\renewcommand{\arraystretch}{1.1}%
\resizebox{0.91\textwidth}{!}
  {\begin{tabular}{|l|c c c|c c c c c c|c c c|}
    \hline
    \multirow{2}{*}{Models} & \multicolumn{3}{c|}{Ubuntu} & \multicolumn{6}{c|}{Douban} & \multicolumn{3}{c|}{E-commerce}\\
      & $R_{10}@1$ & $R_{10}@2$ & $R_{10}@5$ & MAP & MRR & P@1 & $R_{10}@1$ & $R_{10}@2$ & $R_{10}@5$ & $R_{10}@1$ & $R_{10}@2$ & $R_{10}@5$\\
    \hline
    CNN     \cite{kadlec2015improved} & 0.549 & 0.684 & 0.896 & 0.417 & 0.440 & 0.226 & 0.121 & 0.252 & 0.647 & 0.328 & 0.515 & 0.792\\
    LSTM    \cite{kadlec2015improved} & 0.638 & 0.784 & 0.949 & 0.485 & 0.537 & 0.320 & 0.187 & 0.343 & 0.720 & 0.365 & 0.536 & 0.828\\
    BiLSTM  \cite{kadlec2015improved} & 0.630 & 0.780 & 0.944 & 0.479 & 0.514 & 0.313 & 0.184 & 0.330 & 0.716 & 0.365 & 0.536 & 0.825\\
    MV-LSTM       \cite{wan2016match} & 0.653 & 0.804 & 0.946 & 0.498 & 0.538 & 0.348 & 0.202 & 0.351 & 0.710 & 0.412 & 0.591 & 0.857\\ 
    Match-LSTM\cite{wang2016learning} & 0.653 & 0.799 & 0.944 & 0.500 & 0.537 & 0.345 & 0.202 & 0.348 & 0.720 & 0.410 & 0.590 & 0.858\\ 
    \hline
    Multi-View   \cite{zhou2016multi} & 0.662 & 0.801 & 0.951 & 0.505 & 0.543 & 0.342 & 0.202 & 0.350 & 0.729 & 0.421 & 0.601 & 0.861\\
    DL2R           \cite{yan2016dl2r} & 0.626 & 0.783 & 0.944 & 0.488 & 0.527 & 0.330 & 0.193 & 0.342 & 0.705 & 0.399 & 0.571 & 0.842\\
    SMN       \cite{wu2017sequential} & 0.726 & 0.847 & 0.961 & 0.529 & 0.569 & 0.397 & 0.233 & 0.396 & 0.724 & 0.453 & 0.654 & 0.886\\
    DUA           \cite{zhang2018dua} & 0.752 & 0.868 & 0.962 & 0.551 & 0.599 & 0.421 & 0.243 & 0.421 & 0.780 & 0.501 & 0.700 & 0.921\\
    DAM          \cite{zhou2018multi} & 0.767 & 0.874 & 0.969 & 0.550 & 0.601 & 0.427 & 0.254 & 0.410 & 0.757 & 0.526 & 0.727 & 0.933\\
    IoI             \cite{tao2019one} & 0.796 & 0.894 & 0.974 & 0.573 & 0.621 & 0.444 & 0.269 & 0.451 & 0.786 & 0.563 & 0.768 & 0.950\\
    MSN          \cite{yuan2019multi} & 0.800 & 0.899 & 0.978 & 0.587 & 0.632 & 0.470 & 0.295 & 0.452 & 0.788 & 0.606 & 0.770 & 0.937\\
    \hline
    
    
    BERT       \cite{gu2020sabert}    & 0.808 & 0.897 & 0.975 & 0.591 & 0.633 & 0.454 & 0.280 & 0.470 & 0.828 & 0.610 & 0.814 & 0.973\\
    BERT-SS-DA \cite{lu2020improving} & 0.813 & 0.901 & 0.977 & 0.602 & 0.643 & 0.458 & 0.280 & 0.491 & 0.843 & 0.648 & 0.843 & 0.980\\ 
    SA-BERT    \cite{gu2020sabert}    & 0.855 & 0.928 & 0.983 & 0.619 & 0.659 & 0.496 & 0.313 & 0.481 & 0.847 & 0.704 & 0.879 & 0.985\\
    \hline\hline
    BERT (ours)                       & 0.820 & 0.906 & 0.978 & 0.597 & 0.634 & 0.448 & 0.279 & \underline{0.489} & 0.823 & 0.641 & 0.824 & 0.973 \\
    ELECTRA                           & 0.826 & 0.908 & 0.978 & 0.602 & 0.642 & 0.465 & 0.287 & 0.483 & 0.839 & 0.609 & 0.804 & 0.965\\
    \hline
    UMS$_{\text{BERT}}$               & 0.843 & 0.920 & 0.982 & 0.597 & 0.639 & 0.466 & 0.285 & 0.471 & 0.829 & \underline{0.674} & \underline{0.861} & \underline{0.980}\\
    UMS$_{\text{ELECTRA}}$            & \underline{0.854} & \underline{0.929} & \underline{0.984} & \underline{0.608} & \underline{0.650} & \underline{0.472} & \underline{0.291} & 0.488 & \underline{0.845}  & 0.648  & 0.831 & 0.974\\

    \hline\hline
    BERT+                             & 0.862 & 0.935 & 0.987 & 0.609 & 0.645 & 0.463 & 0.290 & \textbf{0.505} & 0.838 & 0.725 & 0.890 & 0.984 \\ 
    ELECTRA+                          & 0.861 & 0.932 & 0.985 & 0.612 & 0.655 & 0.480 & 0.301 & 0.499 & 0.836 & 0.673 & 0.835 & 0.974\\
    \hline
    UMS$_{\text{BERT+}}$              & \textbf{0.875}$^{\dagger}$ & \textbf{0.942}$^{\dagger}$ & \textbf{0.988}$^{\dagger}$ & \textbf{0.625} & \textbf{0.664} & \textbf{0.499} & \textbf{0.318} & 0.482 & \textbf{0.858} & \textbf{0.762} & \textbf{0.905} & \textbf{0.986} \\
    UMS$_{\text{ELECTRA+}}$           & \textbf{0.875} & 0.941 & \textbf{0.988} & 0.623 & 0.663 & 0.492 & 0.307 & 0.501 & 0.851 & 0.707 & 0.853 & 0.974\\

  \hline
  \end{tabular}}
  \caption{Results on Ubuntu, Douban, and E-Commerce datasets. All the evaluation results except ours are cited from published literature \cite{tao2019one,yuan2019multi,gu2020sabert}. The \underline{underlined} numbers mean the best performance for each block and the \textbf{bold} numbers mean state-of-the-art performance for each metric. $\dagger$ denotes statistical significance (p-value $<$ 0.05).} 
  \label{table:quantitative_results}
\end{table*}


\subsection{Evaluation Metrics}
We evaluated our model using several retrieval metrics, following previous research \cite{lowe2015ubuntu,wu2017sequential,zhou2018multi,yuan2019multi}. First, we employ 1 in $n$ recall at $k$, denoted as $R_{n}@k$ ($k=\{1,2,5\}$), which gets 1 when a ground truth is positioned in the $k$ selected list and 0 otherwise. In addition, three other metrics [mean average precision (MAP), mean reciprocal rank (MRR), and precision at one (P@1)] are used especially for Douban and Kakao, as these two datasets may contain more than one positive response among candidates.   

\subsection{Training Details}
We implemented our model\footnote{\begin{footnotesize}\url{https://github.com/taesunwhang/UMS-ResSel}\end{footnotesize}} by using the PyTorch deep learning framework \cite{paszke2019pytorch} based on the open-source code\footnote{\begin{footnotesize}https://github.com/huggingface/transformers\end{footnotesize}} \cite{wolf2019HuggingFacesTS}. As we experimented on three different languages ({\em{i.e.,}} English, Chinese, and Korean), initial checkpoints for BERT and ELECTRA are adapted from several works \cite{devlin2018bert,clark2020electra,cui2020revisiting,lee2020reference}. Specifically, we employ \textit{base} pre-trained models for all languages except for Chinese (the whole-word masking (WWM) strategy is used for Chinese BERT\footnote{\begin{footnotesize}https://github.com/ymcui/Chinese-BERT-wwm\end{footnotesize}}). As ELECTRA for Korean is not available, we do not conduct ELECTRA-based experiments on the Kakao Corpus. All experiments, both post-training and fine-tuning, are run on 4 Tesla V100 GPUs. For fine-tuning, we trained the models with a batch size of 32 using the Adam optimizer with an initial learning rate of 3e-5. The maximum sequence length is set to 512 and $k$ for UMS is set to 5. 

\subsection{Baselines}
\paragraph{Single-turn Matching Models} 
These baselines, including CNN, LSTM, BiLSTM \cite{kadlec2015improved}, MV-LSTM \cite{wan2016match}, and Match-LSTM \cite{wang2016learning}, are based on matching between a dialog context and a response. They construct the dialog context by concatenating utterances and regard it as a long document.

\noindent\textbf{Multi-turn Matching Models}
Multi-View \cite{zhou2016multi} utilize both word- and utterance-level representations. DL2R \cite{yan2016dl2r} reformulates the last utterance with previous utterances in the dialog context. SMN \cite{wu2017sequential} first constructs attention matrices based on word and sequential representations of each utterance and response, and then obtains matching vectors by using CNN. DUA \cite{zhang2018dua} utilizes deep utterance aggregation to form a fine-grained context representation. DAM \cite{zhou2018multi} obtains matching representations of the utterances and response using self- and cross-attention based on Transformer architecture \cite{vaswani2017attention}. IoI \cite{tao2019one} lets utterance--response interaction go deep in a matching model. MSN \cite{yuan2019multi} filters only relevant utterances using a multi-hop selector network.

\noindent\textbf{BERT-based Models} 
Recently, BERT \cite{devlin2018bert} is also applied to response selection, such as vanilla BERT\cite{gu2020sabert}, BERT-SS-DA \cite{lu2020improving}, and SA-BERT \cite{gu2020sabert}. In these models, the dialog context is represented as a long document, as in single-turn matching models. They mainly utilize speaker information of each utterance in the dialog context to extend BERT into a multi-turn fashion.
\section{Results and Discussion}
\subsection{Quantitative Results}
\label{ssec:experiment_results}
Table \ref{table:quantitative_results} lists the quantitative results on Ubuntu, Douban, and E-Commerce datasets. In our experiments, we set two conditions for pre-trained language models. 1) Two different pre-trained language models ({\em{i.e.,}} BERT and ELECTRA) are utilized for fine-tuning. 2) We adapt domain-specific post-training approach (each post-trained model is denoted as BERT+ and ELECTRA+). Based on these initial settings, we explore how effective UMS are for multi-turn response selection. 

For all datasets, models with UMS significantly outperform the previous state-of-the-art methods. Specifically, UMS$_{\text{BERT+}}$ achieves absolute improvements of 2.0\% and 5.8\% in $R_{10}@1$ on Ubuntu and E-Commerce datasets, respectively. For Douban datset, MAP and MRR are considered to be main metrics rather than $R_{10}@1$ because the test set contains more than one ground truth in the candidates. UMS$_{\text{BERT+}}$ achieves absolute improvements of 0.5\% in these metrics.

To evaluate the effectiveness of UMS, we compare the models with UMS and those without them. Since existing BERT-based approaches \cite{lu2020improving, gu2020sabert} reported different performances of BERT, we reimplemented it for a fair comparison with our proposed UMS$_{\text{BERT}}$. The models with UMS consistently show performance improvement regardless of whether language models are post-trained on each corpus or not. For the models without post-training, different results are obtained depending on the dataset. ELECTRA mainly shows better results for the Ubuntu and Douban datasets, while BERT shows better results for the E-Commerce dataset. By contrast, BERT+ achieves the best performance for all corpora in comparison among the models with post-training. We believe that post-training on domain-specific corpus provides the model with more opportunities to learn whether given two dialogs are relevant through NSP; this has the effect of data augmentation.\\
\noindent\textbf{Results on Kakao Corpus} 
We report the evaluation results on the Kakao Corpus in Table \ref{table:kako_performance}. As ELECTRA for Korean is unavailable, we only compare BERT and UMS$_{\text{BERT}}$ for two test splits. \textit{Clean} shows better results than \textit{Web} with respect to all metrics regardless of using UMS. This might be because \textit{Clean} contains fewer grammatical errors and typos that interfere with an accurate understanding of the context. Also, UMS$_{\text{BERT}}$ significantly improves performance compared to the baseline for both split; specifically, it achieves absolute improvements of 5.1\% and 6.8\% in P@1 on \textit{Web} and \textit{Clean}, respectively.
\begin{table}[t]\centering
\small
\begin{adjustbox}{width=0.45\textwidth}

\resizebox{1.0\columnwidth}{!}{
\begin{tabular}{l l c c c c c c}
\toprule
Test Split & Approach & MAP & MRR & $P@1$ & $R_{10}@1$ & $R_{10}@2$ & $R_{10}@5$\\

\midrule
\multirow{2}{*}{Web}   
            & BERT                  & 0.671 & 0.720 & 0.555 & 0.391 & 0.599 & 0.890\\
            & UMS$_{\text{BERT}}$   & \textbf{0.699} & \textbf{0.751} & \textbf{0.606} & \textbf{0.428} & \textbf{0.623} & \textbf{0.911}\\
\midrule
\multirow{2}{*}{Clean}   
            & BERT                   & 0.726 & 0.792 & 0.648 & 0.395 & 0.612 & 0.888\\
            & UMS$_{\text{BERT}}$    & \textbf{0.761} & \textbf{0.834} & \textbf{0.716} & \textbf{0.431} & \textbf{0.663} & \textbf{0.903}\\
\bottomrule
\end{tabular}
}
\end{adjustbox}

\caption{Evaluation Results on Kakao Corpus. }
\label{table:kako_performance}
\end{table}

\subsection{Adversarial Experiment}
\label{ssec:adversarial_experiment}

\begin{table}[t]\centering
\small
\begin{adjustbox}{width=0.4\textwidth}
\resizebox{1.0\columnwidth}{!}{
\begin{tabular}{l l c c c c}
\toprule
\multirow{2}{*}{Approach} & \multirow{2}{*}{Model} & \multicolumn{2}{c}{Original} &\multicolumn{2}{c}{Adversarial} \\
& & $R_{10}@1$ & MRR & $R_{10}@1$ & MRR \\
\midrule

\multirow{5}{*}{Baselines}   
            & BERT      & 0.820 & 0.887 & 0.199 & 0.561\\
            & BERT+     & \textbf{0.862} & \textbf{0.915} & 0.203 & 0.573 \\
            & ELECTRA   & 0.826 & 0.890 & 0.304 & 0.614 \\
            & ELECTRA+  & 0.861 & 0.914 & \textbf{0.329} & \textbf{0.636} \\
            \cmidrule{2-6}
            & Avg       & 0.842 & 0.902 & 0.259 & 0.596\\

\midrule    
\multirow{5}{*}{UMS}
            & BERT      & 0.843 & 0.902 & 0.310 & 0.622\\
            & BERT+     & \textbf{0.875} & \textbf{0.923} & 0.363 & 0.656\\
            & ELECTRA   & 0.854 & 0.910 & 0.397 & 0.668\\
            & ELECTRA+  & \textbf{0.875} & 0.922 & \textbf{0.437} & \textbf{0.692}\\
            \cmidrule{2-6}
            & Avg       & 0.862 & 0.914 & 0.377 & 0.660\\

\bottomrule
\end{tabular}
}
\end{adjustbox}

\caption{Adversarial experimental results on Ubuntu Corpus. All models are evaluated using $R_{10}@1$ and MRR metrics.}
\label{table:adversarial_experiment}
\end{table}

Even though BERT-based models have shown state-of-the-art performance for response selection task, we experiment to know if these models are trained to predict the next utterance properly. Inspired by \citet{jia2017adversarial} and \citet{yuan2019multi}, we design an adversarial experiment to investigate whether language models for response selection are trained properly. 
First, we train the models using the original training set, then evaluate them on either original or adversarial test set. To construct the adversarial test set, we randomly extract an utterance from the dialog context and replace it with one of negative responses among candidates (See Figure~\ref{fig:intro_example}). In adversarial test set, assuming there are $n$ candidates per conversation, a set of candidates consists of a ground truth, an extracted utterance from the dialog context, and $n-$2 negative responses. The extracted utterance is not deleted from the original dialog because it can be crucial for selecting the optimal response.
\begin{figure}[t]\centering
\includegraphics[width=0.4\textwidth]{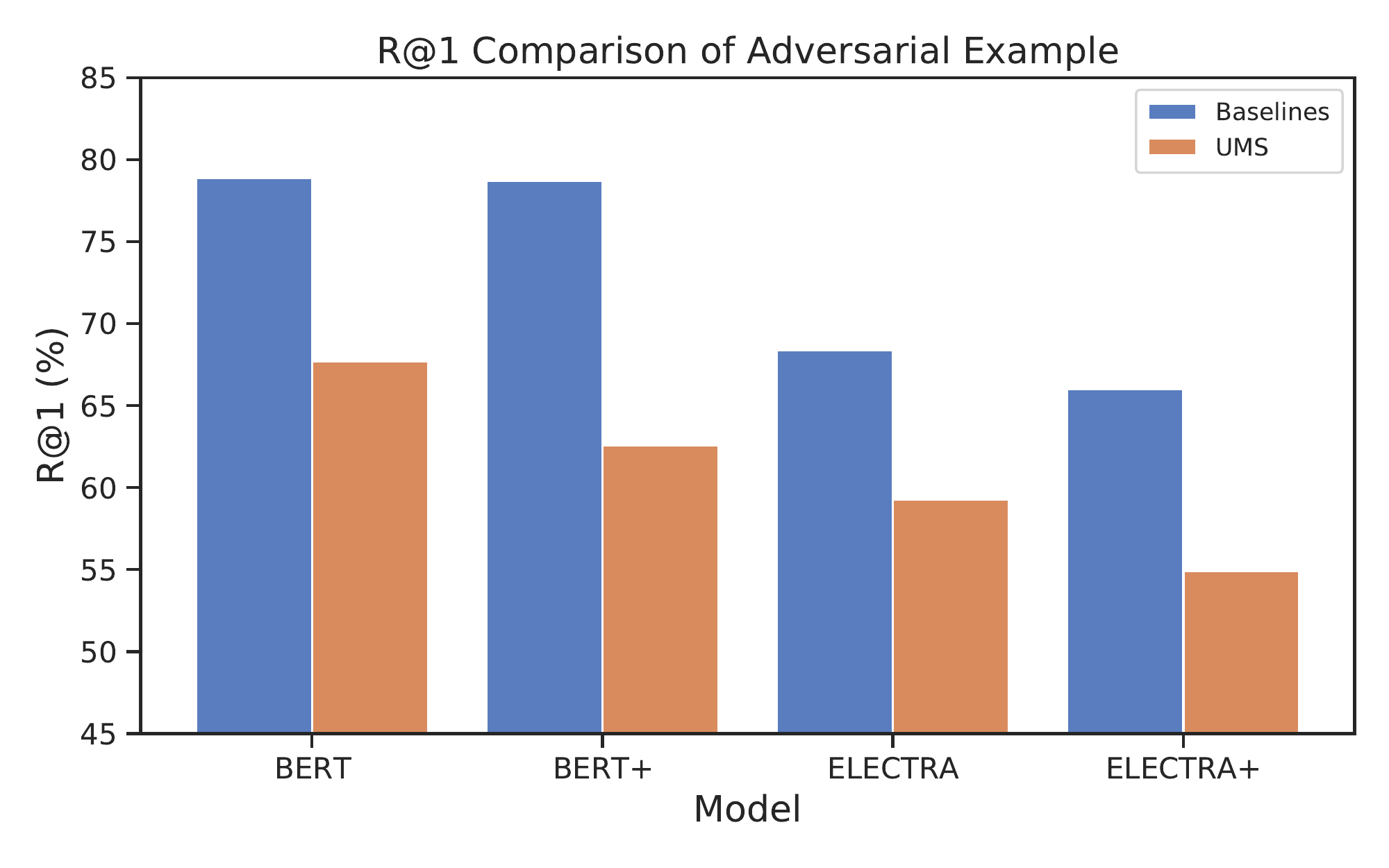}
\caption{$R_{10}@1$ comparison of adversarial example for each model. Lower $R_{10}@1$ means that it is good at predicting the next utterance (ground truth).}
\label{fig:adversarial_recall}
\end{figure}

Table \ref{table:adversarial_experiment} lists the experimental results of BERT(+) and ELECTRA(+) models. We compare the models without UMS and those with, denoted as baselines and UMS, respectively. 
Even though the performances drop significantly in the adversarial set regardless of whether UMS are used or not, we observe that the UMS decline less than baselines. Specifically, $R_{10}@1$ score decreases by 58\% and 48\% on average for baselines and UMS, respectively. It is also encouraging that UMS show an absolute improvement of 12\% with respect to $R_{10}@1$ on the adversarial set compared to the 2\% improvement on the original set (See Table \ref{table:adversarial_experiment}). In addition, while baselines tend to drop in performance on the adversarial set as training progresses, the performance of UMS shows a tendency to increase significantly. Hence, it is reasonable to assume that our UMS are robust to adversarial examples and are good at \textit{in-domain} classification.

Figure \ref{fig:adversarial_recall} describes the performance of each model, ranking adversarial example ({\em{i.e.,}} randomly sampled utterance from the conversation) as the most likely response. While BERT- and ELECTRA-based models show similar performance on the original set, ELECTRA-based models outperform BERT-based models with significant margins (a gap of 10\%) on the adversarial set regardless of whether they are trained from post-trained checkpoints. For example, different patterns of the evaluation results between BERT+ and ELECTRA are observed according to the test sets (original : BERT+ $>$ ELECTRA, adversarial : BERT+ $<$ ELECTRA). We have two perspectives on these results: 1) \textit{Next sentence prediction} in BERT overfits the model to predict semantically relevant sentence rather than the next sentence. 2) As ELECTRA is trained through \textit{replaced token detection} in which the model learns to discriminate between real input tokens and replacements generated from small \textit{masked language model}, it is more effective in representing contextual information from the sequence.

\subsection{Ablation Study}
We performed ablation studies on the Ubuntu Corpus to investigate which auxiliary tasks are more crucial for response selection. As shown in Table \ref{table:ablation_study}, we explored the impact of each auxiliary task by constructing all combinations of possible subsets. Based on the observations of using only one auxiliary task ({\em{i.e.,}} 3 $>$ 2 $\approx$ 4) and two tasks ({\em{i.e.,}} 5 $\approx$ 7 $>$ 6), we obtained the results, DEL $>$ INS $\approx$ SRCH, with respect to the importance of manipulation strategy. As DEL consists of an input sequence that contains an irrelevant utterance to the original dialog context, it may be more advantageous for learning to distinguish dialog consistency and coherence than INS and SRCH. We obtain the best results when all the auxiliary tasks are trained simultaneously with the response selection criterion.


\begin{table}[t]\centering
\small
\begin{adjustbox}{width=0.4\textwidth}

\resizebox{1.0\columnwidth}{!}{
\begin{tabular}{p{0.02\textwidth} l c c c c c c c }
\toprule
& Auxiliary Tasks & $R_{10}@1$ & $R_{10}@2$ & $R_{10}@5$ & MRR \\
\midrule 
1 & None & 0.826 & 0.908 & 0.978 & 0.890\\
\midrule
2 & INS & 0.836 & 0.917 & 0.980 & 0.897\\
3 & DEL & 0.848 & 0.924 & 0.983 & 0.905\\
4 & SRCH & 0.834 & 0.915 & 0.981 & 0.896\\
\midrule
5 & INS + DEL  & 0.853 & 0.927 & 0.984 & 0.909\\
6 & INS + SRCH & 0.841 & 0.920 & 0.982 & 0.901\\
7 & DEL + SRCH & 0.852 & 0.927 & 0.983 & 0.908\\
\midrule

8 & INS + DEL + SRCH & \textbf{0.854} & \textbf{0.929} & \textbf{0.984} & \textbf{0.910}\\
\bottomrule

\end{tabular}
}
\end{adjustbox}

\caption{Ablation Study on Ubuntu Corpus. We choose ELECTRA as the baseline in this analysis. INS, DEL, and SRCH denote that the model trained with utterance insertion, deletion, and search, respectively.}
\label{table:ablation_study}
\end{table}


\begin{figure}[t]\centering
\includegraphics[width=0.43\textwidth]{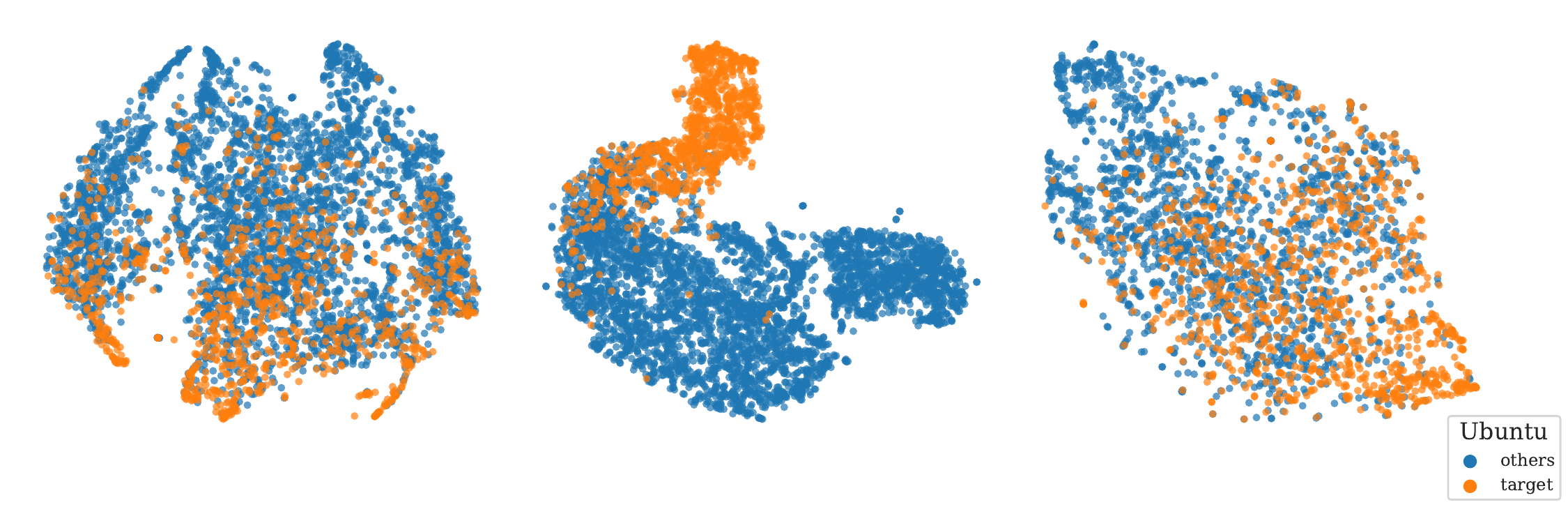}
\includegraphics[width=0.43\textwidth]{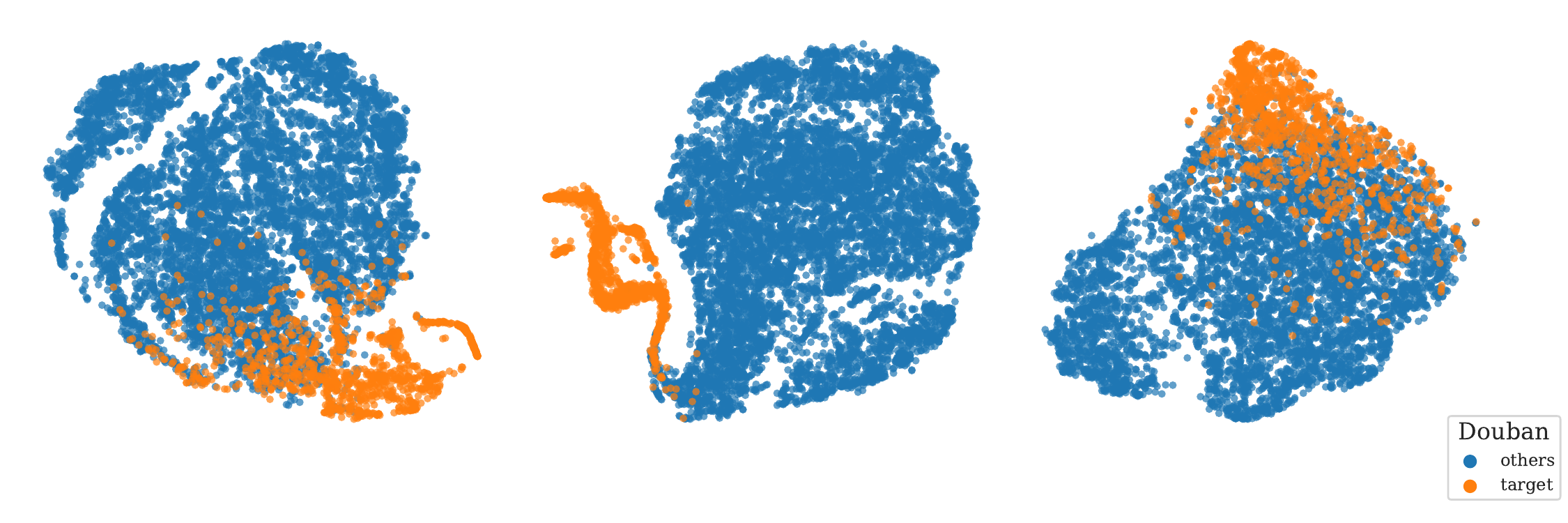}
\includegraphics[width=0.43\textwidth]{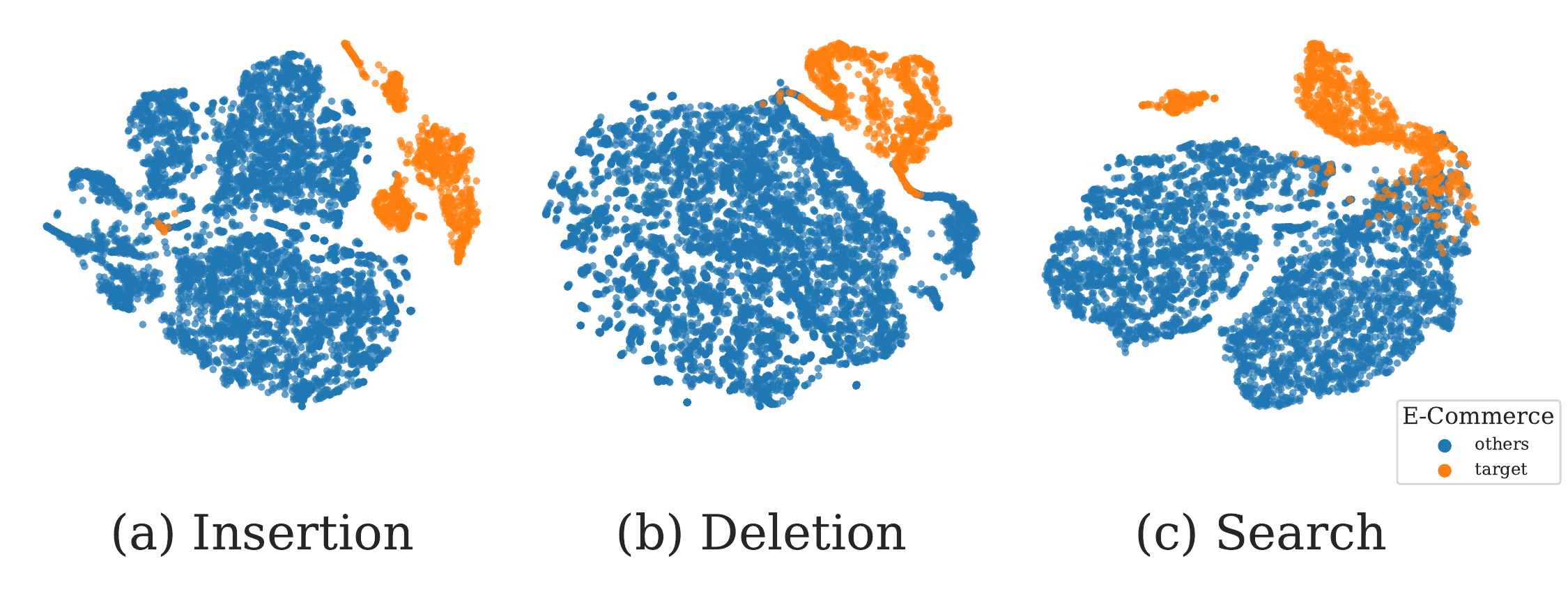}

\caption{t-SNE embeddings of UMS$_\text{BERT+}$ output representations for each special token in UMS ({\em{i.e.,}} \ins\,, \del\,, and \srch\,). All embeddings are sampled from test sets of each dataset.}
\label{fig:t_sne}
\end{figure}

\subsection{Visualization}
As shown in Figure \ref{fig:t_sne}, we visualize the output representations of special tokens learned by our proposed UMS through t-SNE embeddings. Scatter plots colored in orange represent target tokens ({\em{i.e.,}}\ins$_t$,\,\del$_t$, and \srch$_t$) and those in blue represent the rest of tokens. All representations are extracted from test sets of three datasets (Ubuntu, Douban, and E-Commerce) in this analysis. Overall, the results show that UMS$_\text{BERT+}$ effectively learns dialog coherence for all datasets. In the case of Ubuntu dataset, insertion and search tasks tend to be less clustered than that of the other two datsets. As many utterances in the Ubuntu dataset mainly consist of many technical terminologies that may cause structural ambiguity, tasks constructed within the same dialog are difficult to be performed. By contrast, the model can easily learn discourse structures on open-domain datasets such as Douban and E-Commerce.
\section{Related Work}
\noindent \textbf{Multi-turn Response Selection} Early approaches to response selection focused on single-turn response selection~\cite{wang2013dataset,hu2014convolutional,wang2015syntax}. Recently, multi-turn response selection has received more attention by researchers. \citet{lowe2015ubuntu} proposed dual encoder architecture which uses an RNN-based models to match the dialog and response. \citet{zhou2016multi} proposed the multi-view model that encodes dialog context and response both on the word-level and utterance-level. However, these models have limitations in fully reflecting the relationship between dialog and response. 
To alleviate this, \citet{wu2017sequential} proposed the sequential matching network that utilizes matching matrices to match each utterance with a response. As self-attention \cite{vaswani2017attention} mechanism has been proved its effectiveness, it is applied in subsequent works \cite{zhou2018multi,tao2019multi,tao2019one}. \citet{yuan2019multi} recently pointed out that previous approaches construct dialog representation with abundant information but noisy, which deteriorates the performance. They proposed an effective history filtering technique to avoid using excessive history information. 

Most recently, many researches based on pre-trained language models including BERT \cite{devlin2018bert} and RoBERTa \cite{liu2019roberta} are proposed. Generally, most models formulate the response selection task as a dialog-response binary classification task. \citet{whang2019domain} first applied BERT for multi-turn response selection and obtained state-of-the-art results through further training BERT on domain-specific corpus. Subsequent researches \cite{lu2020improving,gu2020sabert} focused on modeling speaker information and showed its effectiveness in response retrieval. \citet{humeau2019poly} investigated the trade-off relationship between model complexity and computation efficiency in the language models. They proposed poly-encoders that ensure fast inference speed, even though the performance is slightly lower than that of the cross-encoder.

\noindent\textbf{Self-supervised Learning} Self-supervised learning has been explored in various pre-trained language models \cite{devlin2018bert,clark2020electra,lewis2020bart, joshi2020spanbert} and is also applied in several NLP downstream tasks, such as summarization \cite{wang2019self_summ}, disfluency detection \cite{wang2020multi}, and response generation \cite{zhao2020learning}. Existing works in dialog modeling \cite{wu2019self,mehri2019pretraining,liang2020beyond} mainly focused on building enhanced dialog representations through self-supervised learning. Although the methods proposed in \citet{wu2019self} and \citet{liang2020beyond} effectively learn to rank coherent dialog higher than corrupted ones, but they have limitations in identifying the utterance that actually caused the inconsistency. Our strategy is different in that it learns to find which utterance is replaced from the full dialog. By doing so, our model can learn which utterance does not suit the conversation, which makes the model learn not only to discriminate semantic differences but also to build coherent dialog. The method proposed in \citet{mehri2019pretraining} is somewhat similar to our deletion task, but they directly use the utterance representation to build the loss. We hypothesize that this is the reason behind the inconsistent improvements in \citet{mehri2019pretraining}, where in some downstream tasks the auxiliary task was actually harmful. On the other hand, our approach introduces special tokens that is different from the \cls\, token used in downstream tasks. Our results show that this approach consistently leads to improvements.


\section{Conclusion}
In this paper, we pointed out the limitations of existing works based on pre-trained language models such as BERT in retrieval-based multi-turn dialog systems. To address these, we proposed highly effective utterance manipulation strategies (UMS) for multi-turn response selection. The UMS are fully applied in self-supervised manner and can be easily incorporated into existing models. We obtained new state-of-the-art results on multiple public benchmark datasets ({\em{i.e.,}} Ubuntu, Douban, and E-Commerce) and significantly improved results on Korean open-domain dialog corpus. For the future work, we plan to develop a response selection model which is more robust to adversarial examples by designing various adversarial objectives. 
\section{Acknowledgments}
This work was partly supported by Institute for Information \& communications Technology Planning \& Evaluation(IITP) grant funded by the Korea government(MSIT) (No. 1711117120, A Neural-Symbolic Model for Knowledge Acquisition and Inference Techniques) and funded by the Korean National Police Agency (No. PR09-01-000-20, Pol-Bot Development for Conversational Police Knowledge Services)

\bibliography{aaai2021}

\begin{thebibliography}{40}
\providecommand{\natexlab}[1]{#1}
\providecommand{\url}[1]{\texttt{#1}}
\providecommand{\urlprefix}{URL }
\expandafter\ifx\csname urlstyle\endcsname\relax
  \providecommand{\doi}[1]{doi:\discretionary{}{}{}#1}\else
  \providecommand{\doi}{doi:\discretionary{}{}{}\begingroup
  \urlstyle{rm}\Url}\fi

\bibitem[{Clark et~al.(2020)Clark, Luong, Le, and Manning}]{clark2020electra}
Clark, K.; Luong, M.-T.; Le, Q.~V.; and Manning, C.~D. 2020.
\newblock ELECTRA: Pre-training Text Encoders as Discriminators Rather Than
  Generators.
\newblock In \emph{International Conference on Learning Representations}.

\bibitem[{Cui et~al.(2020)Cui, Che, Liu, Qin, Wang, and Hu}]{cui2020revisiting}
Cui, Y.; Che, W.; Liu, T.; Qin, B.; Wang, S.; and Hu, G. 2020.
\newblock Revisiting Pre-Trained Models for Chinese Natural Language
  Processing.
\newblock \emph{arXiv preprint arXiv:2004.13922} .

\bibitem[{Devlin et~al.(2019)Devlin, Chang, Lee, and
  Toutanova}]{devlin2018bert}
Devlin, J.; Chang, M.-W.; Lee, K.; and Toutanova, K. 2019.
\newblock {BERT}: Pre-training of Deep Bidirectional Transformers for Language
  Understanding.
\newblock In \emph{Proceedings of the 2019 Conference of the North {A}merican
  Chapter of the Association for Computational Linguistics: Human Language
  Technologies, Volume 1 (Long and Short Papers)}, 4171--4186.

\bibitem[{Gu et~al.(2020)Gu, Li, Liu, Ling, Su, Wei, and Zhu}]{gu2020sabert}
Gu, J.-C.; Li, T.; Liu, Q.; Ling, Z.-H.; Su, Z.; Wei, S.; and Zhu, X. 2020.
\newblock Speaker-Aware BERT for Multi-Turn Response Selection in
  Retrieval-Based Chatbots.
\newblock In \emph{Proceedings of the 29th ACM International Conference on
  Information and Knowledge Management}.

\bibitem[{Hu et~al.(2014)Hu, Lu, Li, and Chen}]{hu2014convolutional}
Hu, B.; Lu, Z.; Li, H.; and Chen, Q. 2014.
\newblock Convolutional neural network architectures for matching natural
  language sentences.
\newblock In \emph{Advances in neural information processing systems},
  2042--2050.

\bibitem[{Humeau et~al.(2020)Humeau, Shuster, Lachaux, and
  Weston}]{humeau2019poly}
Humeau, S.; Shuster, K.; Lachaux, M.-A.; and Weston, J. 2020.
\newblock Poly-encoders: Architectures and Pre-training Strategies for Fast and
  Accurate Multi-sentence Scoring.
\newblock In \emph{International Conference on Learning Representations}.

\bibitem[{Jia and Liang(2017)}]{jia2017adversarial}
Jia, R.; and Liang, P. 2017.
\newblock Adversarial Examples for Evaluating Reading Comprehension Systems.
\newblock In \emph{Proceedings of the 2017 Conference on Empirical Methods in
  Natural Language Processing}, 2021--2031.

\bibitem[{Joshi et~al.(2020)Joshi, Chen, Liu, Weld, Zettlemoyer, and
  Levy}]{joshi2020spanbert}
Joshi, M.; Chen, D.; Liu, Y.; Weld, D.~S.; Zettlemoyer, L.; and Levy, O. 2020.
\newblock Spanbert: Improving pre-training by representing and predicting
  spans.
\newblock \emph{Transactions of the Association for Computational Linguistics}
  8: 64--77.

\bibitem[{Kadlec, Schmid, and Kleindienst(2015)}]{kadlec2015improved}
Kadlec, R.; Schmid, M.; and Kleindienst, J. 2015.
\newblock Improved deep learning baselines for ubuntu corpus dialogs.
\newblock \emph{arXiv preprint arXiv:1510.03753} .

\bibitem[{Lan et~al.(2020)Lan, Chen, Goodman, Gimpel, Sharma, and
  Soricut}]{lan2020albert}
Lan, Z.; Chen, M.; Goodman, S.; Gimpel, K.; Sharma, P.; and Soricut, R. 2020.
\newblock ALBERT: A Lite BERT for Self-supervised Learning of Language
  Representations.
\newblock In \emph{International Conference on Learning Representations}.

\bibitem[{Lee et~al.(2020)Lee, Shin, Whang, Cho, Ko, Lee, Kim, and
  Jo}]{lee2020reference}
Lee, D.; Shin, M.~C.; Whang, T.; Cho, S.; Ko, B.; Lee, D.; Kim, E.; and Jo, J.
  2020.
\newblock Reference and Document Aware Semantic Evaluation Methods for {K}orean
  Language Summarization.
\newblock In \emph{Proceedings of the 28th International Conference on
  Computational Linguistics}, 5604--5616.

\bibitem[{Lewis et~al.(2020)Lewis, Liu, Goyal, Ghazvininejad, Mohamed, Levy,
  Stoyanov, and Zettlemoyer}]{lewis2020bart}
Lewis, M.; Liu, Y.; Goyal, N.; Ghazvininejad, M.; Mohamed, A.; Levy, O.;
  Stoyanov, V.; and Zettlemoyer, L. 2020.
\newblock {BART}: Denoising Sequence-to-Sequence Pre-training for Natural
  Language Generation, Translation, and Comprehension.
\newblock In \emph{Proceedings of the 58th Annual Meeting of the Association
  for Computational Linguistics}, 7871--7880.

\bibitem[{Liang, Zou, and Yu(2020)}]{liang2020beyond}
Liang, W.; Zou, J.; and Yu, Z. 2020.
\newblock Beyond User Self-Reported {L}ikert Scale Ratings: A Comparison Model
  for Automatic Dialog Evaluation.
\newblock In \emph{Proceedings of the 58th Annual Meeting of the Association
  for Computational Linguistics}, 1363--1374.

\bibitem[{Liu et~al.(2019)Liu, Ott, Goyal, Du, Joshi, Chen, Levy, Lewis,
  Zettlemoyer, and Stoyanov}]{liu2019roberta}
Liu, Y.; Ott, M.; Goyal, N.; Du, J.; Joshi, M.; Chen, D.; Levy, O.; Lewis, M.;
  Zettlemoyer, L.; and Stoyanov, V. 2019.
\newblock Roberta: A robustly optimized bert pretraining approach.
\newblock \emph{arXiv preprint arXiv:1907.11692} .

\bibitem[{Lowe et~al.(2015)Lowe, Pow, Serban, and Pineau}]{lowe2015ubuntu}
Lowe, R.; Pow, N.; Serban, I.; and Pineau, J. 2015.
\newblock The {U}buntu Dialogue Corpus: A Large Dataset for Research in
  Unstructured Multi-Turn Dialogue Systems.
\newblock In \emph{Proceedings of the 16th Annual Meeting of the Special
  Interest Group on Discourse and Dialogue}, 285--294.

\bibitem[{Lu et~al.(2020)Lu, Ren, Ren, Liu, and Xu}]{lu2020improving}
Lu, J.; Ren, X.; Ren, Y.; Liu, A.; and Xu, Z. 2020.
\newblock Improving Contextual Language Models for Response Retrieval in
  Multi-Turn Conversation.
\newblock In \emph{Proceedings of the 43rd International ACM SIGIR Conference
  on Research and Development in Information Retrieval}, 1805--1808.

\bibitem[{Mehri et~al.(2019)Mehri, Razumovskaia, Zhao, and
  Eskenazi}]{mehri2019pretraining}
Mehri, S.; Razumovskaia, E.; Zhao, T.; and Eskenazi, M. 2019.
\newblock Pretraining Methods for Dialog Context Representation Learning.
\newblock In \emph{Proceedings of the 57th Annual Meeting of the Association
  for Computational Linguistics}, 3836--3845.

\bibitem[{Paszke et~al.(2019)Paszke, Gross, Massa, Lerer, Bradbury, Chanan,
  Killeen, Lin, Gimelshein, Antiga et~al.}]{paszke2019pytorch}
Paszke, A.; Gross, S.; Massa, F.; Lerer, A.; Bradbury, J.; Chanan, G.; Killeen,
  T.; Lin, Z.; Gimelshein, N.; Antiga, L.; et~al. 2019.
\newblock Pytorch: An imperative style, high-performance deep learning library.
\newblock In \emph{Advances in neural information processing systems},
  8026--8037.

\bibitem[{Tao et~al.(2019{\natexlab{a}})Tao, Wu, Xu, Hu, Zhao, and
  Yan}]{tao2019multi}
Tao, C.; Wu, W.; Xu, C.; Hu, W.; Zhao, D.; and Yan, R. 2019{\natexlab{a}}.
\newblock Multi-representation fusion network for multi-turn response selection
  in retrieval-based chatbots.
\newblock In \emph{Proceedings of the Twelfth ACM International Conference on
  Web Search and Data Mining}, 267--275.

\bibitem[{Tao et~al.(2019{\natexlab{b}})Tao, Wu, Xu, Hu, Zhao, and
  Yan}]{tao2019one}
Tao, C.; Wu, W.; Xu, C.; Hu, W.; Zhao, D.; and Yan, R. 2019{\natexlab{b}}.
\newblock One Time of Interaction May Not Be Enough: Go Deep with an
  Interaction-over-Interaction Network for Response Selection in Dialogues.
\newblock In \emph{Proceedings of the 57th Annual Meeting of the Association
  for Computational Linguistics}, 1--11.

\bibitem[{Vaswani et~al.(2017)Vaswani, Shazeer, Parmar, Uszkoreit, Jones,
  Gomez, Kaiser, and Polosukhin}]{vaswani2017attention}
Vaswani, A.; Shazeer, N.; Parmar, N.; Uszkoreit, J.; Jones, L.; Gomez, A.~N.;
  Kaiser, {\L}.; and Polosukhin, I. 2017.
\newblock Attention is all you need.
\newblock In \emph{Proceedings of the Advances in Neural Information Processing
  Systems}, 5998--6008.

\bibitem[{Wan et~al.(2016)Wan, Lan, Xu, Guo, Pang, and Cheng}]{wan2016match}
Wan, S.; Lan, Y.; Xu, J.; Guo, J.; Pang, L.; and Cheng, X. 2016.
\newblock Match-SRNN: Modeling the Recursive Matching Structure with Spatial
  RNN.
\newblock In \emph{Proceedings of the 25th International Joint Conference on
  Artificial Intelligence}, 2922–2928.

\bibitem[{Wang et~al.(2019{\natexlab{a}})Wang, Pruksachatkun, Nangia, Singh,
  Michael, Hill, Levy, and Bowman}]{wang2019superglue}
Wang, A.; Pruksachatkun, Y.; Nangia, N.; Singh, A.; Michael, J.; Hill, F.;
  Levy, O.; and Bowman, S. 2019{\natexlab{a}}.
\newblock Superglue: A stickier benchmark for general-purpose language
  understanding systems.
\newblock In \emph{Proceedings of the Advances in Neural Information Processing
  Systems}, 3266--3280.

\bibitem[{Wang et~al.(2013)Wang, Lu, Li, and Chen}]{wang2013dataset}
Wang, H.; Lu, Z.; Li, H.; and Chen, E. 2013.
\newblock A dataset for research on short-text conversations.
\newblock In \emph{Proceedings of the 2013 Conference on Empirical Methods in
  Natural Language Processing}, 935--945.

\bibitem[{Wang et~al.(2019{\natexlab{b}})Wang, Wang, Xiong, Yu, Guo, Chang, and
  Wang}]{wang2019self_summ}
Wang, H.; Wang, X.; Xiong, W.; Yu, M.; Guo, X.; Chang, S.; and Wang, W.~Y.
  2019{\natexlab{b}}.
\newblock Self-Supervised Learning for Contextualized Extractive Summarization.
\newblock In \emph{Proceedings of the 57th Annual Meeting of the Association
  for Computational Linguistics}, 2221--2227.

\bibitem[{Wang et~al.(2015)Wang, Lu, Li, and Liu}]{wang2015syntax}
Wang, M.; Lu, Z.; Li, H.; and Liu, Q. 2015.
\newblock Syntax-Based Deep Matching of Short Texts.
\newblock In \emph{Proceedings of the 24th International Joint Conference on
  Artificial Intelligence}, 1354–1361.

\bibitem[{Wang et~al.(2020)Wang, Che, Liu, Qin, Liu, and Wang}]{wang2020multi}
Wang, S.; Che, W.; Liu, Q.; Qin, P.; Liu, T.; and Wang, W.~Y. 2020.
\newblock Multi-task self-supervised learning for disfluency detection.
\newblock In \emph{Proceedings of the AAAI Conference on Artificial
  Intelligence}, 9193--9200.

\bibitem[{Wang and Jiang(2016)}]{wang2016learning}
Wang, S.; and Jiang, J. 2016.
\newblock Learning Natural Language Inference with {LSTM}.
\newblock In \emph{Proceedings of the 2016 Conference of the North {A}merican
  Chapter of the Association for Computational Linguistics: Human Language
  Technologies}, 1442--1451.

\bibitem[{Whang et~al.(2020)Whang, Lee, Lee, Yang, Oh, and
  Lim}]{whang2019domain}
Whang, T.; Lee, D.; Lee, C.; Yang, K.; Oh, D.; and Lim, H. 2020.
\newblock An Effective Domain Adaptive Post-Training Method for BERT in
  Response Selection.
\newblock In \emph{Proc. Interspeech 2020}, 1585--1589.

\bibitem[{Wolf et~al.(2019)Wolf, Debut, Sanh, Chaumond, Delangue, Moi, Cistac,
  Rault, Louf, Funtowicz, and Brew}]{wolf2019HuggingFacesTS}
Wolf, T.; Debut, L.; Sanh, V.; Chaumond, J.; Delangue, C.; Moi, A.; Cistac, P.;
  Rault, T.; Louf, R.; Funtowicz, M.; and Brew, J. 2019.
\newblock HuggingFace's Transformers: State-of-the-art Natural Language
  Processing.
\newblock \emph{arXiv preprint arXiv:1910.03771} .

\bibitem[{Wu, Wang, and Wang(2019)}]{wu2019self}
Wu, J.; Wang, X.; and Wang, W.~Y. 2019.
\newblock Self-Supervised Dialogue Learning.
\newblock In \emph{Proceedings of the 57th Annual Meeting of the Association
  for Computational Linguistics}, 3857--3867.

\bibitem[{Wu et~al.(2017)Wu, Wu, Xing, Zhou, and Li}]{wu2017sequential}
Wu, Y.; Wu, W.; Xing, C.; Zhou, M.; and Li, Z. 2017.
\newblock Sequential Matching Network: A New Architecture for Multi-turn
  Response Selection in Retrieval-Based Chatbots.
\newblock In \emph{Proceedings of the 55th Annual Meeting of the Association
  for Computational Linguistics (Volume 1: Long Papers)}, 496--505.

\bibitem[{Xu et~al.(2019)Xu, Liu, Shu, and Philip}]{xu2019bert}
Xu, H.; Liu, B.; Shu, L.; and Philip, S.~Y. 2019.
\newblock BERT Post-Training for Review Reading Comprehension and Aspect-based
  Sentiment Analysis.
\newblock In \emph{Proceedings of the 2019 Conference of the North American
  Chapter of the Association for Computational Linguistics: Human Language
  Technologies, Volume 1 (Long and Short Papers)}, 2324--2335.

\bibitem[{Xu et~al.(2017)Xu, Liu, Wang, Sun, and Wang}]{xu2017incorporating}
Xu, Z.; Liu, B.; Wang, B.; Sun, C.; and Wang, X. 2017.
\newblock Incorporating loose-structured knowledge into conversation modeling
  via recall-gate LSTM.
\newblock In \emph{2017 International Joint Conference on Neural Networks
  (IJCNN)}, 3506--3513.

\bibitem[{Yan, Song, and Wu(2016)}]{yan2016dl2r}
Yan, R.; Song, Y.; and Wu, H. 2016.
\newblock Learning to Respond with Deep Neural Networks for Retrieval-Based
  Human-Computer Conversation System.
\newblock In \emph{Proceedings of the 39th International ACM SIGIR Conference
  on Research and Development in Information Retrieval}, 55--64.

\bibitem[{Yuan et~al.(2019)Yuan, Zhou, Li, Lv, Zhu, Han, and
  Hu}]{yuan2019multi}
Yuan, C.; Zhou, W.; Li, M.; Lv, S.; Zhu, F.; Han, J.; and Hu, S. 2019.
\newblock Multi-hop Selector Network for Multi-turn Response Selection in
  Retrieval-based Chatbots.
\newblock In \emph{Proceedings of the 2019 Conference on Empirical Methods in
  Natural Language Processing and the 9th International Joint Conference on
  Natural Language Processing (EMNLP-IJCNLP)}, 111--120.

\bibitem[{Zhang et~al.(2018)Zhang, Li, Zhu, Zhao, and Liu}]{zhang2018dua}
Zhang, Z.; Li, J.; Zhu, P.; Zhao, H.; and Liu, G. 2018.
\newblock Modeling Multi-turn Conversation with Deep Utterance Aggregation.
\newblock In \emph{Proceedings of the 27th International Conference on
  Computational Linguistics}, 3740--3752.

\bibitem[{Zhao, Xu, and Wu(2020)}]{zhao2020learning}
Zhao, Y.; Xu, C.; and Wu, W. 2020.
\newblock Learning a Simple and Effective Model for Multi-turn Response
  Generation with Auxiliary Tasks.
\newblock In \emph{Proceedings of the 2020 Conference on Empirical Methods in
  Natural Language Processing (EMNLP)}, 3472--3483.

\bibitem[{Zhou et~al.(2016)Zhou, Dong, Wu, Zhao, Yu, Tian, Liu, and
  Yan}]{zhou2016multi}
Zhou, X.; Dong, D.; Wu, H.; Zhao, S.; Yu, D.; Tian, H.; Liu, X.; and Yan, R.
  2016.
\newblock Multi-view Response Selection for Human-Computer Conversation.
\newblock In \emph{Proceedings of the 2016 Conference on Empirical Methods in
  Natural Language Processing}, 372--381.

\bibitem[{Zhou et~al.(2018)Zhou, Li, Dong, Liu, Chen, Zhao, Yu, and
  Wu}]{zhou2018multi}
Zhou, X.; Li, L.; Dong, D.; Liu, Y.; Chen, Y.; Zhao, W.~X.; Yu, D.; and Wu, H.
  2018.
\newblock Multi-Turn Response Selection for Chatbots with Deep Attention
  Matching Network.
\newblock In \emph{Proceedings of the 56th Annual Meeting of the Association
  for Computational Linguistics (Volume 1: Long Papers)}, 1118--1127.

\end{thebibliography}

\end{document}